\documentclass{article}
\usepackage{spconf,amsmath,amsfonts,graphicx,verbatim,booktabs}
\usepackage{amsmath,bm}
\usepackage{stfloats}
\usepackage{setspace}
\usepackage[nobreak]{cite}
\usepackage{multirow}
\usepackage{subcaption}
\usepackage{array}

\title{Interpretable Visual Question Answering \\Referring to Outside Knowledge}
\name{He Zhu$^{\dag}$, Ren Togo$^{\dag\dag}$, Takahiro Ogawa$^{\dag\dag}$ and Miki Haseyama$^{\dag\dag}$\thanks{This work was partly supported by JSPS KAKENHI Grant Number JP21H03456.}}
\address{
    $^{\dag}$Graduate School of Information Science and Technology, Hokkaido University, Japan\\
    $^{\dag\dag}$Faculty of Information Science and Technology, Hokkaido University, Japan\\
    E-mail: \{zhu, togo, ogawa, mhaseyama\}@lmd.ist.hokudai.ac.jp
    }

\begin{document}
\ninept
\maketitle
%

\begin{abstract}
We present a novel multimodal interpretable VQA model that can answer the question more accurately and generate diverse explanations. Although researchers have proposed several methods that can generate human-readable and fine-grained natural language sentences to explain a model's decision, these methods have focused solely on the information in the image. Ideally, the model should refer to various information inside and outside the image to correctly generate explanations, just as we use background knowledge daily. The proposed method incorporates information from outside knowledge and multiple image captions to increase the diversity of information available to the model. The contribution of this paper is to construct an interpretable visual question answering model using multimodal inputs to improve the rationality of generated results. Experimental results show that our model can outperform state-of-the-art methods regarding answer accuracy and explanation rationality.
\end{abstract}
\begin{keywords}
Visual question answering, interpretable machine learning, outside knowledge learning, natural language explanations. 
\end{keywords}
\section{Introduction}
\label{sec:intro}
Visual Question Answering (VQA)~\cite{antol2015vqa} has been widely used in various scenarios such as defect detection~\cite{czimmermann2020visual} and medical diagnosis~\cite{lin2021medical}. Since these applications are closely related to the safety and property of our life, there are high requirements for the reliability of the VQA methods.
Although most present methods have demonstrated exemplary performance, they are considered black boxes, as their internal logic and workings are concealed from users, which cannot meet the reliability requirements.
It is considered that providing a reasonable explanation for the prediction process can become one of the solutions to prove the model's reliability. Therefore, new regulations have been created to establish standards for verifying mandatory decisions, which increases the demand for methods' interpretability~\cite{molnar2020interpretable}.

Most conventional interpretable VQA models use a heat map to represent the different attention parts of the model in an image. Although this representation can explain the prediction somewhat, it can still be abstract to users.
To solve this problem, Park et al.~\cite{park2018multimodal} and Wu et al.~\cite{wu2018faithful} have proposed interpretable VQA models to generate explanations with human-readable and detailed natural language sentences.
These methods require creating another model that generates explanations for the VQA model. Due to the generation of explanations and answers being independent, it is difficult to control whether the generated explanation conforms to the logic used by the VQA model.
Recently, pre-trained large-scale models such as the Bidirectional Encoder Representations from Transformers (BERT)~\cite{devlin2018bert}, GPT-2~\cite{radford2019language}, and the Contrastive Language-Image Pre-training (CLIP)~\cite{radford2021learning} have been developed rapidly, which makes it possible to generate natural language more correctly.
Focusing on the deficiency of the previous models, Sammani et al.~\cite{sammani2022nlx}, Marasovic et al.~\cite{RVT}, and Kayser et al.~\cite{DBLP:journals/corr/abs-2105-03761} provided intrinsic interpretability models, which used a single model to generate answers and explanations simultaneously. 

Although these methods ensure the underlying logic's consistency, the generated results' performance is still flawed.
\begin{figure}[t]
\centering  
\includegraphics[width=8cm]{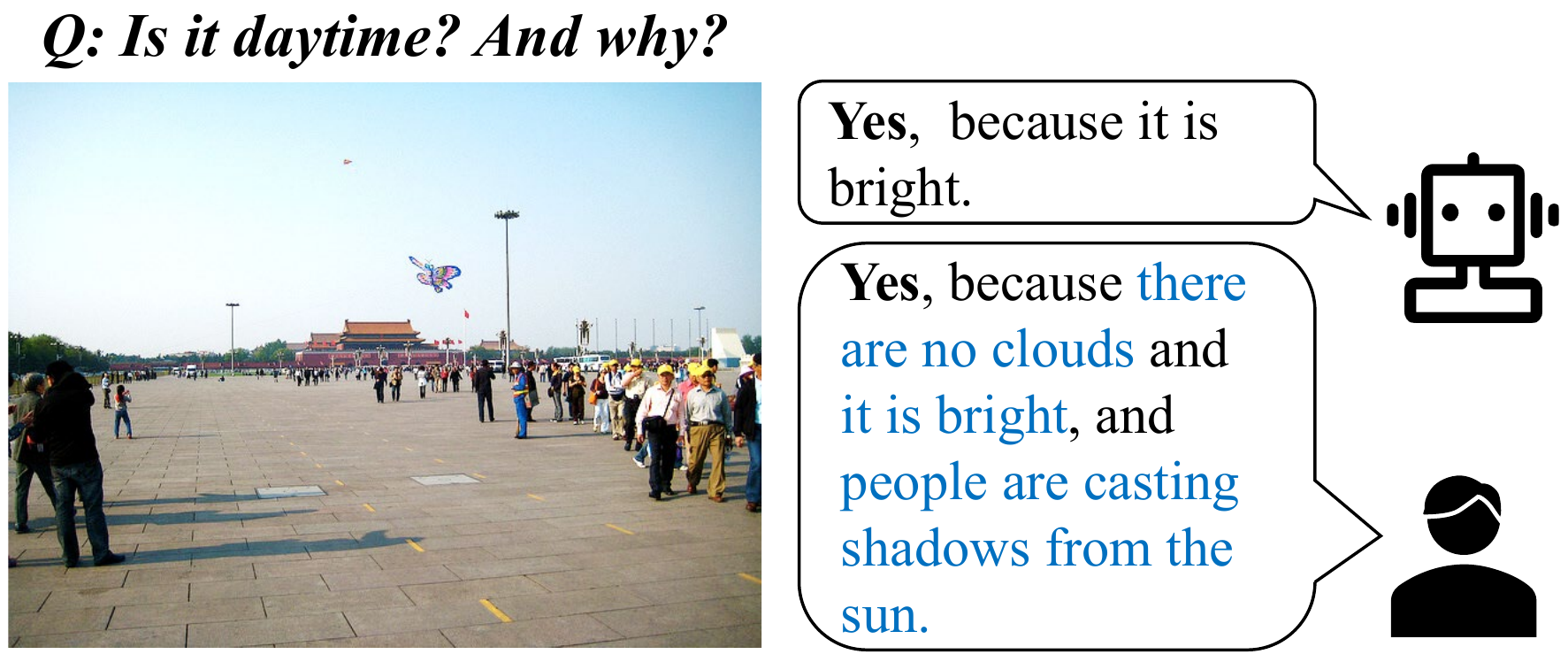}
\caption{Limitation of the existing interpretable VQA methods.}
\label{fig:intro}
\vspace{-2em}
\end{figure}
As shown in Fig.~\ref{fig:intro}, human-provided explanations are more complex and informative than previous interpretable methods. The reason is that answering a question in our daily lives requires drawing on a wide range of knowledge and experience beyond just what we can see and recognize in an image. For example, people have a wealth of background knowledge in and outside the image and usually refer to them to explain an answer. Hence, simply referring to visual information to answer questions is still insufficient.
As Zhu et al.~\cite{He2022_10} demonstrated, incorporating image captions as an additional modality is a valuable extension. However, their approach of introducing metadata to the model's input does not align with the VQA task's requirement of only image input. Additionally, the utilization of a single caption suffers from the limitation that many captions are irrelevant to the problem, and the provided information only pertains to the image's interior. As a result, this strategy fails to address the issue of single-source information and leads to a monotonous interpretation.
Outside knowledge is always considered to contain information outside the image. Thus, to solve the above informativeness problem, outside knowledge should be introduced to improve the model's performance.
Moreover, we need to use caption generation models to generate multiple captions to introduce more helpful information without changing the model inputs.

This paper proposes a novel multimodal interpretable VQA model that can refer to various information while generating results by introducing outside knowledge and multiple image captions. 
Concretely, we newly generate the image's caption and use the generated content to obtain the most relevant outside knowledge from the wikidata~\cite{vrandevcic2014wikidata}. 
In the proposed multimodal method, we use a joint vector consisting of the features of the image, its caption, and outside knowledge to resolve the conflict between different domains.
By leveraging the novel joint vector, our model can draw on a variety of information sources to produce more rational explanations and more accurate answers.
Finally, we summarize our contributions of this paper as follows.
\begin{itemize}
\item We newly introduce the outside knowledge and multiple image captions to the multimodal interpretable VQA model to solve the informativeness problem.
\item Our method outperforms other state-of-the-art models, as evidenced by experimental results that demonstrate more accurate answers and more comprehensive and reasonable explanations.
\end{itemize}
\section{NATURAL LANGUAGE EXPLANATION GENERATION
VIA OUTSIDE-KNOWLEDGE AND MULTIPLE CAPTIONS}
\label{sec:format}
\begin{figure*}[t]
\centering   
\includegraphics[width=18cm]{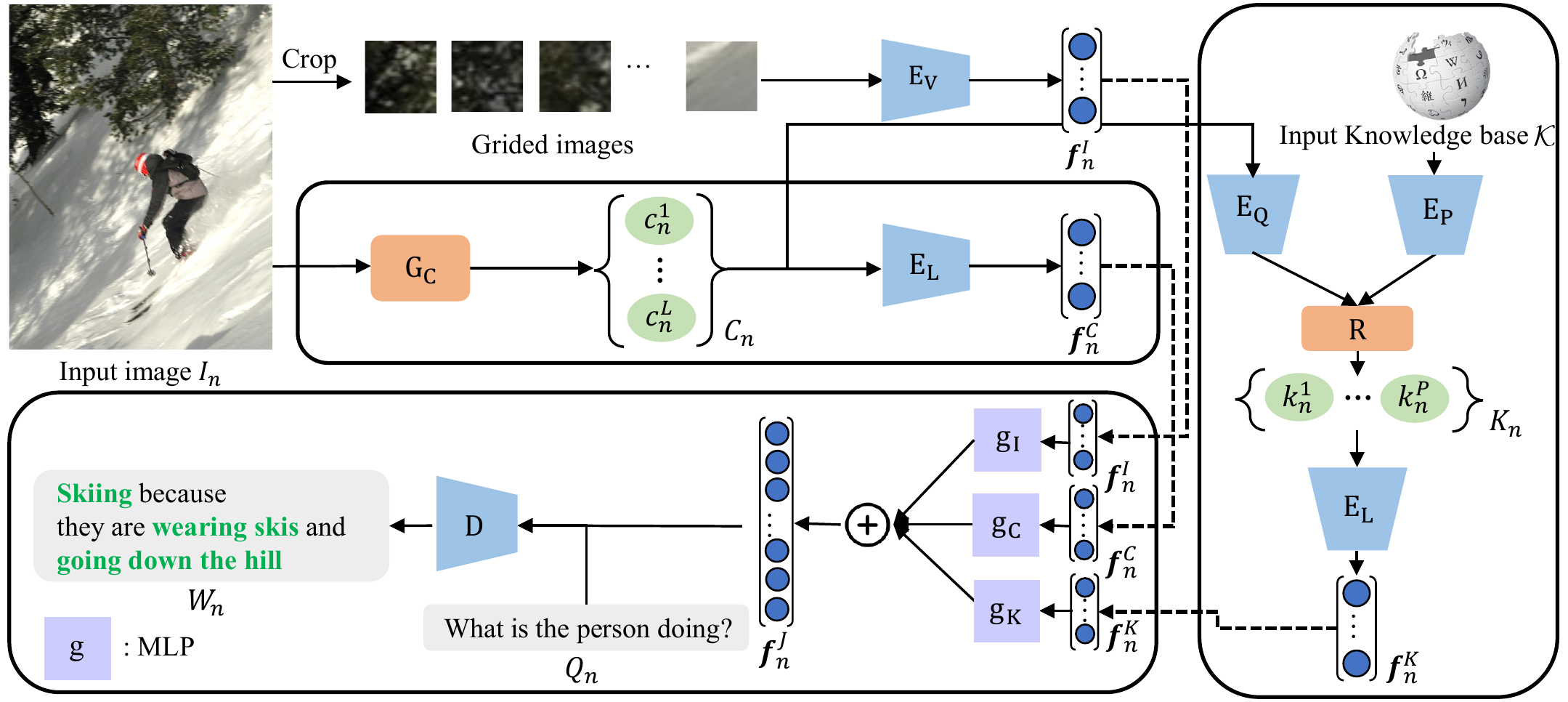}
\caption{An overview of our model. The model answers the input question $Q_n$ and generates a corresponding explanation with human-friendly natural language sentence $W_n$ based on the input image $I_n$. We newly introduce the image caption set $C_n$ and outside knowledge set $K_n$ into the proposed model to refer to various information during the generation process.}
\vspace{-2em}
\label{fig:overview}
\end{figure*}
We provide an overview of our model in Fig.~\ref{fig:overview}. Compared with the previous models with separate answering and explaining modules, we propose a VQA model that simultaneously answers questions and generates explanations.
To improve the rationality of the generated explanations and the answer accuracy, instead of only using the input image, our model newly refers to the information contained in the retrieved outside knowledge and multiple image captions.
\subsection{Image Feature Extraction}
\begin{table*}[t]
    \centering
    \caption{Evaluation of model accuracy against other state-of-the-art models. The metrics for explanation evaluate the accuracy of the word level. Acc. means the accuracy of the model answering the question, and the higher score indicates better model performance. ``Ours w/o OK" and ``Ours w/o C" represent only introducing caption and outside knowledge, respectively. ``Ours via BERT" represents that BERT is used as the language encoder.}
    		\begin{tabular}{lccccccccc}\hline
    		    &\multicolumn{8}{c}{Explanation}&\multicolumn{1}{c}{Answer}\\
    		    \cmidrule(lr){2-9} \cmidrule(lr){10-10} 
    			&BLEU-1&BLEU-2&BLEU-3&BLEU-4&ROUGE-L&METEOR&CIDEr&SPICE&Acc.\\\hline\hline
    			NLX-GPT~\cite{sammani2022nlx}&64.0&48.6&36.4&27.2&50.7&22.6&104.7&21.6&67.1\\\hline
    			E-ViL~\cite{DBLP:journals/corr/abs-2105-03761}&52.6&36.6&24.9&17.2&40.3&19.0&51.8&15.7&56.8\\\hline
    			QA-only~\cite{RVT}&59.3&44.0&32.3&23.9&47.4&20.6&91.8&17.9&63.4\\\hline
    			RVT~\cite{RVT}&59.4&43.4&31.1&22.3&46.6&20.1&84.4&17.3&62.7\\\hline\hline
    			\textbf{Ours w/o OK}&64.0&49.1&37.2&\textbf{28.3}&51.3&23.1&\textbf{109.8}&21.3&68.5\\
    			\textbf{Ours w/o C}&63.9&48.7&36.7&27.6&51.1&22.6&106.1&21.4&67.4\\
    			\textbf{Ours (via BERT)}&64.3&49.0&37.1&28.1&51.3&\textbf{23.2}&108.0&21.3&67.3\\\hline
    			\textbf{Ours}&\textbf{65.2}&\textbf{49.8}&\textbf{37.4}&\textbf{28.3}&\textbf{51.6}&\textbf{23.2}&108.1&\textbf{21.6}&\textbf{69.5}\\\hline
    		\end{tabular}
    		\label{tab:result acc}
\vspace{-0.5em}
\end{table*}
For each of $N$ training data instances, we denote it as $n$-th instance ($n \in 1, 2, ..., N$). Our method takes an image $I_n$, a question $Q_n$, and a ground truth sentence $G_n$. Then we generate an output sentence $W_n$ of length $M$, where $w_m$ refers to $m$-th word in the sentence, which contains both the answer and explanation.

We clip the image before extracting the feature as $N_{\rm grid} \times N_{\rm grid}$ ($N_{\rm grid}$ being the size of the grid) small pieces, which are represented as $I_n = \lbrace p_1, p_2, ..., p_{N_{\rm grid} \times N_{\rm grid}}\rbrace$. Then we use a transformer-based vision encoder ${\rm E_V}$ to extract the feature of the input image.
The image feature $\bm{f}_n^I$ can be calculated as follows:
\begin{equation}
\bm{f}_n^I = {\rm E_V}(\lbrace p_1, p_2, ..., p_{N_{\rm grid} \times N_{\rm grid}}\rbrace).
\end{equation}
Through the encoder $\rm E_V$ with an attention structure, we can extract image features and capture the relationship between regions in $\bm{f}_n^I$, which is the crucial information to answer a question and generate an explanation.
\subsection{Introduction of Image Caption and Outside Knowledge}
A knowledge base containing various types of knowledge is needed to retrieve the outside knowledge of images.
We set our sights on Wikipedia, which collects a large amount of structured data and has become a resource of enormous value with potential applications.
We finally chose a recently proposed subset from Wikidata~\cite{vrandevcic2014wikidata} following KAT~\cite{gui2021kat} as our outside knowledge set $\mathcal{K}$ in this paper.

Since the image caption and the outside knowledge are natural language contents, we use a language encoder ${\rm E_L}(\cdot)$ to extract their features in the model.
We first use a pre-trained image captioning model ${\rm G_C}(\cdot)$ to generate a caption set $C_n=\{c_n^1,...,c_n^L\}$ containing $L$ captions from the $I_n$. The features of each caption in $C_n$ are extracted using ${\rm E_L}(\cdot)$, and then the obtained features are joined into a caption feature vector $\bm{f}_n^C$ by summing.
Next, we use a retrieval model ${\rm R}(\cdot,\cdot)$ and the query $C_n$ to retrieve the knowledge item set $K_n=\{k_n^1,...,k_n^P\}$ containing $P$ retrieved knowledge items from $\mathcal{K}$.

We use a query encoder ${\rm E_Q}(\cdot)$ and a passage encoder ${\rm E_P}(\cdot)$ to encode $C_n$ and $\mathcal{K}$, respectively, and then use the extracted features to get the retrieved outside knowledge $K_n$ by using the retrieval model ${\rm R}(\cdot,\cdot)$, which can be calculated as follows:
\begin{equation}
K_n = {\rm R}({\rm E_Q}(C_n), {\rm E_P}(\mathcal{K})).
\end{equation}
To introduce the outside knowledge into the final prediction process, we extract the outside knowledge feature vector $\bm{f}_n^{K}$ using the ${\rm E_L}(\cdot)$ in the same way as extracting $\bm{f}_n^C$.

In this way, we extract the feature vectors of the image caption and outside knowledge containing the internal and external information of the image, which will then be joined with the vision vector in the final prediction process.
\subsection{Generation of Answer and Explanation} 
We use a joint vector to expand the information because feature fusion is an essential step in multimodal tasks~\cite{dai2021attentional}. Because of the characteristics of our vision and language encoder, the image and language features are mapped to latent shared spaces. The concatenating operation allows us to use their features best and make our model contain much information.
However, even if the extracted features are in a shared space, the simple concatenation between the different modalities will still lead to difficulties in model training. Thus, we use Multi-layer Perceptron (MLP) layers to better fuse the features with considering the combination effect and learning efficiency.
First, three different MLPs, ${\rm g_I}$, ${\rm g_C}$ and ${\rm g_K}$, are used to process different reference information separately. Then we use the concatenation operation to generate the joint vector $\bm{f}_n^J$ as follows:
\begin{equation}
    \bm{f}_n^J = <{\rm g_C}(\bm{f}_n^{C}), {\rm g_K}(\bm{f}_n^{K}), {\rm g_I}(\bm{f}_n^I)>.
\end{equation}
Through this feature fusion process, we can join the feature vectors of the different modalities to preserve information as much as possible.

The output of our model is a sentence $W_n$, which contains the answer to question $Q_n$ and the corresponding explanation. We use a decoder ${\rm D}$ to decode $\bm{f}_n^J$ to generate the context:$\{question\} + \{answer\} + because + \{explanation\}$ as follows:
\begin{equation}
W_n = {\rm D}(\bm{f}_n^J, Q_n),
\end{equation}
where $Q_n$ is the input, and the decoder generates the answer and the explanation $W_n$. To train the model, we employ the cross-entropy loss and minimize the negative log-likelihood, which can be computed as
\begin{equation}
\mathcal{L} = -\sum_{m=1}^N {\rm log}\,{p_\theta}(w_t|\bm{w}_{<m}).
\end{equation}
The probability mass function is represented by ${p_\theta}(\cdot)$, where $\theta$ is a parameter of the model distribution. The term $\bm{w}_{<m}$ refers to the words preceding $w_t$. By minimizing the loss $\mathcal{L}$, our model can produce accurate explanations that closely resemble the ground truth.
\section{EXPERIMENTS}
\label{sec:pagestyle}
\begin{figure*}[t]
\centering
\includegraphics[width=18cm]{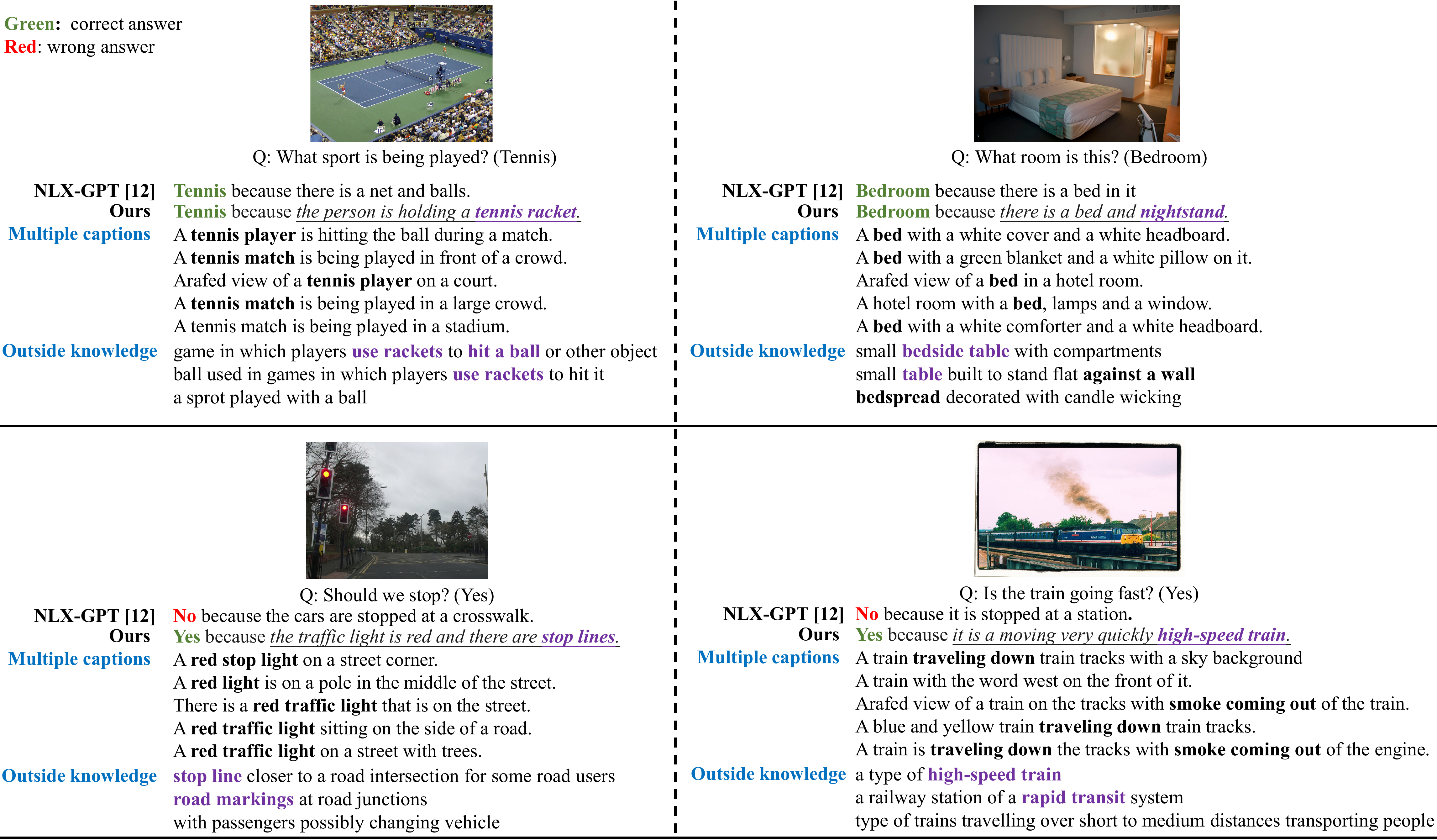}
\caption{Samples of the comparison between our model and NLX-GPT. The results show that our model can generate a more reasonable explanation based on the useful information contained in the image caption and the outside knowledge. We offer the generated captions and outside knowledge of each image, and the correct answers are in parentheses. We chose four representative examples according to whether the answer given by the model is correct or not.}
\label{fig:res}
\end{figure*}
\subsection{Experimental Settings}
$\mathbf{Dataset}$.
For our experiments, we utilized the VQA-X dataset~\cite{park2018multimodal}, an extension of the VQA-v2 dataset~\cite{antol2015vqa}, which included explanations for each answer. 
The dataset comprises 33k QA pairs from 28k images sourced from the COCO2014 dataset~\cite{lin2014microsoft}. 
For the dataset division, we used all the images in the COCO2014 training set with 29k QA pairs as our training set. We divided the COCO2014 validation set into our validation set and test set according to the proportion of 3:4, which contains 1.5k QA pairs and 2k QA pairs, respectively.

Compared to traditional vision models with specific tasks such as image classification and image segmentation, for the vision encoder, we only rely on their primary network function to output simple grid features rather than their time-consuming top-down features. 
Thus, to better adapt to the Vision \& Language task, we used the CLIP based on the structure of the vision transformer as the vision encoder.
The CLIP makes the fusion of vision and language features easier by encoding them in the same hidden space.
After experimentally verifying two of the most widely used pre-trained language models, BERT and CLIP-text, for the language encoder, we selected the CLIP-text model, which can better match the vision and language features. 
We used five different models to generate multiple captions containing various information, including GIT-large~\cite{wang2022git} fine-tuned on COCO, GIT-large~\cite{wang2022git} fine-tuned on TextCaps, BLIP~\cite{li2022blip}, CoCa~\cite{yu2022coca} and BLIP-2~\cite{BLIP2}.
In the outside knowledge retrieval, we used the same query and passage encoder as Dense Passage Retrieval~\cite{karpukhin2020dense} and the Faiss~\cite{johnson2019billion} to realize the retrieval process.

All the MLPs in our model have three layers and a 128 hidden size, and we set the caption number $L$ to 5 and the outside knowledge item number $P$ to 3.
During the training, we resized all images to 224 $\times$ 224 pixels and added random flips. We trained our models for 30 epochs using a batch size of 32 and a learning rate of $2 \times 10^{-5}$, which was decreased to $10^{-5}$.

$\mathbf{Comparison\ Methods}$. 
The comparison methods answer the question and generate natural language explanations based only on the input image. The NLX-GPT~\cite{sammani2022nlx} and E-ViL~\cite{DBLP:journals/corr/abs-2105-03761} have separate answering process and explanation process, and RVT and QA-only~\cite{RVT} combine the two processes.

$\mathbf{Evaluation\ Metrics}$. 
We used the accuracy to evaluate the generated answers and used the following common language modeling evaluation metrics: BLEU-\textit{n} (\textit{n} being 1 to 4)~\cite{papineni2002bleu}, METEOR~\cite{banerjee2005meteor}, ROUGE~\cite{lin2003automatic}, SPICE~\cite{anderson2016spice} and CIDEr~\cite{vedantam2015cider}. All the scores of the language metrics were computed by the publicly available code provided by Chen et al.~\cite{DBLP:journals/corr/ChenFLVGDZ15}.

\subsection{Experimental Results}
As shown in Table~\ref{tab:result acc}, the obtained results prove that our model performs better than the comparison methods.
Since introducing multiple references, we conducted ablation experiments to verify the validity of each reference information and the language encoder selection. Specifically, we experimented separately with models that only introduce outside knowledge (``Ours w/o C'') and caption (``Ours w/o OK''). Experimental results show that models that introduce captions or outside knowledge as additional references perform better than models that just refer to images. However, the model that uses captions and outside knowledge (``Ours'') performs best. These results prove that if the model has more information to refer to, it can generate answers and their corresponding explanations more correctly.
The results of ``Ours (via BERT)'' also show that using the same language encoder CLIP-text model as the vision encoder can better align language and vision features and generate better results than the BERT.

We show the qualitative comparison between our model with the state-of-the-art method, NLX-GPT, in Fig.~\ref{fig:res}. We have bolded in black and purple some critical information in the caption and outside knowledge that is helpful for the results, respectively. We can see that this information is used for the generated answers and explanations.
As shown in the example in the first row, with the same correct answer, we refer to more information. Thus, our approach yields a more detailed and reasonable explanation.
Meanwhile, in the second row, our method correctly answers questions that the comparison model cannot answer because of the reference to the additional outside knowledge and image caption.
\section{CONCLUSION}
\label{sec:typestyle}
By incorporating multimodal reference information and advanced interpretability techniques, our VQA model generates highly precise explanations. The proposed method can solve the previous informativeness problem by additionally referring to the image caption and outside knowledge. The questions answered by our model were also more accurate. The qualitative and quantitative assessment results demonstrate that our approach for generating answers and explanations surpasses several state-of-the-art methods.



\newpage
\small
\bibliographystyle{IEEEbib}
\bibliography{strings,refs}

\begin{thebibliography}{10}

\bibitem{antol2015vqa}
Stanislaw Antol, Aishwarya Agrawal, Jiasen Lu, Margaret Mitchell, Dhruv Batra,
  C~Lawrence Zitnick, and Devi Parikh,
\newblock ``{VQA}: Visual question answering,''
\newblock in {\em Proceedings of the IEEE International Conference on Computer
  Vision}, 2015, pp. 2425--2433.

\bibitem{czimmermann2020visual}
Tam{\'a}s Czimmermann and others.,
\newblock ``Visual-based defect detection and classification approaches for
  industrial applications—a survey,''
\newblock {\em Sensors}, p. 1459, 2020.

\bibitem{lin2021medical}
Zhihong Lin, Donghao Zhang, Qingyi Tac, Danli Shi, Gholamreza Haffari, Qi~Wu,
  Mingguang He, and Zongyuan Ge,
\newblock ``Medical visual question answering: A survey,''
\newblock {\em arXiv preprint arXiv:2111.10056}, 2021.

\bibitem{molnar2020interpretable}
Christoph Molnar,
\newblock {\em Interpretable machine learning: A Guide for Making Black Box
  Models Explainable},
\newblock Available online:
  https://christophm.github.io/interpretable-ml-book/, 2020.

\bibitem{park2018multimodal}
Park et~al.,
\newblock ``Multimodal explanations: Justifying decisions and pointing to the
  evidence,''
\newblock in {\em Proceedings of the IEEE Conference on Conference Vision and
  Pattern Recognition}, 2018, pp. 8779--8788.

\bibitem{wu2018faithful}
Jialin Wu and Raymond Mooney,
\newblock ``Faithful multimodal explanation for visual question answering,''
\newblock in {\em Proceedings of the ACL Workshop Blackbox NLP: Analyzing and
  Interpreting Neural Networks for NLP}, 2019, pp. 103--112.

\bibitem{devlin2018bert}
Jacob Devlin, Ming-Wei Chang, Kenton Lee, and Kristina Toutanova,
\newblock ``Bert: Pre-training of deep bidirectional transformers for language
  understanding,''
\newblock {\em arXiv preprint arXiv:1810.04805}, 2018.

\bibitem{radford2019language}
Alec Radford, Jeffrey Wu, Rewon Child, David Luan, Dario Amodei, Ilya
  Sutskever, et~al.,
\newblock ``Language models are unsupervised multitask learners,''
\newblock {\em OpenAI {B}log}, vol. 1, no. 8, pp. 9, 2019.

\bibitem{radford2021learning}
Alec Radford, Jong~Wook Kim, Chris Hallacy, Ramesh, et~al.,
\newblock ``Learning transferable visual models from natural language
  supervision,''
\newblock in {\em Proceedings of the International Conference on Machine
  Learning}, 2021, pp. 8748--8763.

\bibitem{sammani2022nlx}
Fawaz Sammani, Tanmoy Mukherjee, and Nikos Deligiannis,
\newblock ``{NLX-GPT}: A model for natural language explanations in vision and
  vision-language tasks,''
\newblock in {\em Proceedings of the IEEE Conference on Computer Vision and
  Pattern Recognition}, 2022, pp. 8322--8332.

\bibitem{RVT}
Ana Marasovi{\'c} et~al.,
\newblock ``Natural language rationales with full-stack visual reasoning: From
  pixels to semantic frames to commonsense graphs,''
\newblock in {\em Proceedings of the Findings of the Association for
  Computational Linguistics: EMNLP}, 2020, pp. 2810--2829.

\bibitem{DBLP:journals/corr/abs-2105-03761}
Kayser et~al.,
\newblock ``{E-ViL}: A dataset and benchmark for natural language explanations
  in vision-language tasks,''
\newblock in {\em Proceedings of the IEEE International Conference on Computer
  Vision}, 2021, pp. 1244--1254.

\bibitem{He2022_10}
He~Zhu, Ren Togo, Takahiro Ogawa, and Miki Haseyama,
\newblock ``A multimodal interpretable visual question answering model
  introducing image caption processor,''
\newblock in {\em Proceedings of the IEEE Global Conference on Consumer
  Electronics}, 2022, pp. 805--806.

\bibitem{vrandevcic2014wikidata}
Denny Vrande{\v{c}}i{\'c} and Markus Kr{\"o}tzsch,
\newblock ``Wikidata: a free collaborative knowledgebase,''
\newblock {\em Communications of the ACM}, pp. 78--85, 2014.

\bibitem{gui2021kat}
Gui et~al.,
\newblock ``{KAT}: A knowledge augmented transformer for vision-and-language,''
\newblock {\em arXiv preprint arXiv:2112.08614}, 2021.

\bibitem{dai2021attentional}
Yimian Dai, Fabian Gieseke, Stefan Oehmcke, Yiquan Wu, and Kobus Barnard,
\newblock ``Attentional feature fusion,''
\newblock in {\em Proceedings of the IEEE Winter Conference on Applications of
  Computer Vision}, 2021, pp. 3560--3569.

\bibitem{lin2014microsoft}
Tsung-Yi Lin, Michael Maire, Serge Belongie, Hays, et~al.,
\newblock ``Microsoft {COCO}: Common objects in context,''
\newblock in {\em Proceedings of the European Conference on Computer Vision},
  2014, pp. 740--755.

\bibitem{wang2022git}
Jianfeng Wang et~al.,
\newblock ``{GIT}: A generative image-to-text transformer for vision and
  language,''
\newblock {\em Transactions on Machine Learning Research}, 2022.

\bibitem{li2022blip}
Junnan Li, Dongxu Li, Caiming Xiong, and Steven Hoi,
\newblock ``Blip: Bootstrapping language-image pre-training for unified
  vision-language understanding and generation,''
\newblock in {\em Proceedings of the International Conference on Machine
  Learning}. PMLR, 2022, pp. 12888--12900.

\bibitem{yu2022coca}
Jiahui Yu, Zirui Wang, Vijay Vasudevan, Legg Yeung, Mojtaba Seyedhosseini, and
  Yonghui Wu,
\newblock ``Coca: Contrastive captioners are image-text foundation models,''
\newblock {\em Transactions on Machine Learning Research}, 2022.

\bibitem{BLIP2}
Junnan Li, Dongxu Li, Silvio Savarese, and Steven Hoi,
\newblock ``Blip-2: Bootstrapping language-image pre-training with frozen image
  encoders and large language models,''
\newblock {\em arXiv preprint arXiv:2301.12597}, 2023.

\bibitem{karpukhin2020dense}
Vladimir Karpukhin, Barlas O{\u{g}}uz, Sewon Min, Patrick Lewis, Ledell Wu,
  Sergey Edunov, Danqi Chen, and Wen-tau Yih,
\newblock ``Dense passage retrieval for open-domain question answering,''
\newblock {\em arXiv preprint arXiv:2004.04906}, 2020.

\bibitem{johnson2019billion}
Jeff Johnson, Matthijs Douze, and Herv{\'e} J{\'e}gou,
\newblock ``Billion-scale similarity search with {GPUs},''
\newblock {\em IEEE Transactions on Big Data}, pp. 535--547, 2019.

\bibitem{papineni2002bleu}
Papineni et~al.,
\newblock ``{BLEU}: a method for automatic evaluation of machine translation,''
\newblock in {\em Proceedings of the Annual Meeting of the Association for
  Computational Linguistics}, 2002, pp. 311--318.

\bibitem{banerjee2005meteor}
Satanjeev Banerjee and Alon Lavie,
\newblock ``{METEOR}: An automatic metric for mt evaluation with improved
  correlation with human judgments,''
\newblock in {\em Proceedings of the ACL workshop on Intrinsic and Extrinsic
  Evaluation Measures for Machine Translation and/or Summarization}, 2005, pp.
  65--72.

\bibitem{lin2003automatic}
Chin-Yew Lin and Eduard Hovy,
\newblock ``Automatic evaluation of summaries using n-gram co-occurrence
  statistics,''
\newblock in {\em Proceedings of the Human Language Technology Conference of
  the North American Chapter of the Association for Computational linguistics},
  2003, pp. 150--157.

\bibitem{anderson2016spice}
Peter Anderson, Basura Fernando, Mark Johnson, and Stephen Gould,
\newblock ``{SPICE}: Semantic propositional image caption evaluation,''
\newblock in {\em Proceedings of the European Conference on Conference Vision},
  2016, pp. 382--398.

\bibitem{vedantam2015cider}
Ramakrishna Vedantam, C~Lawrence~Zitnick, and Devi Parikh,
\newblock ``{CIDE}r: Consensus-based image description evaluation,''
\newblock in {\em Proceedings of the IEEE Conference on Computer Vision and
  Pattern Recognition}, 2015, pp. 4566--4575.

\bibitem{DBLP:journals/corr/ChenFLVGDZ15}
Xinlei Chen, Hao Fang, Tsung-Yi Lin, Ramakrishna Vedantam, Saurabh Gupta, Piotr
  Doll{\'a}r, and C~Lawrence Zitnick,
\newblock ``Microsoft {COCO} captions: Data collection and evaluation server,''
\newblock {\em arXiv preprint arXiv:1504.00325}, 2015.

\end{thebibliography}
\end{document}